\documentclass[letterpaper]{article} % DO NOT CHANGE THIS
\usepackage{aaai2026}  % DO NOT CHANGE THIS
\usepackage{times}  % DO NOT CHANGE THIS
\usepackage{helvet}  % DO NOT CHANGE THIS
\usepackage{courier}  % DO NOT CHANGE THIS
\usepackage[hyphens]{url}  % DO NOT CHANGE THIS
\usepackage{graphicx} % DO NOT CHANGE THIS
\usepackage{natbib}  % DO NOT CHANGE THIS AND DO NOT ADD ANY OPTIONS TO IT
\usepackage{caption} % DO NOT CHANGE THIS AND DO NOT ADD ANY OPTIONS TO IT
\usepackage{algorithm}
\usepackage{algorithmic}

\usepackage{amsmath,amsfonts,amsthm,nicefrac,amssymb}
\usepackage{mathtools}
\usepackage{bm}
\usepackage{pifont}
\usepackage{graphicx}
\usepackage[hyphens]{url}
\usepackage{xcolor}
\usepackage{soul}
\usepackage{natbib}
\usepackage{booktabs}
\usepackage{multicol}
\usepackage{multirow}
\usepackage{subcaption}

\usepackage{cleveref}

\DeclareMathOperator*{\rank}{\mathrm{rank}}
\DeclareMathOperator*{\softmax}{\mathrm{SoftMax}}

\newcommand{\defeq}{:=}
\newcommand{\expect}{\mathbb{E}}

\renewcommand{\mid}{\,\vert\,}

\newcommand{\fim}{\mathcal{I}}
\newcommand{\T}{\top}
\newcommand{\calB}{\mathcal{B}}

\newcommand{\calV}{\mathcal{V}}
\newcommand{\calE}{\mathcal{E}}
\newcommand{\calG}{\mathcal{G}}
\newcommand{\calT}{\mathcal{T}}

\newcommand{\calY}{\mathcal{Y}}
\newcommand{\calR}{\mathcal{R}}

\newcommand{\calQ}{\mathcal{Q}}
\newcommand{\calM}{\mathcal{M}}

\newcommand{\tildewl}{\widetilde{\bm{W}}^l}
\newcommand{\grad}{\bigtriangledown}

\theoremstyle{plain}
\newtheorem{theorem}{Theorem}
\newtheorem{proposition}[theorem]{Proposition}
\newtheorem*{remark}{Remark}

\title{Unbiased Online Curvature Approximation for Regularized Graph Continual Learning}
\author {
    Jie Yin\textsuperscript{\rm 1},
    Ke Sun\textsuperscript{\rm 2},
    Han Wu\textsuperscript{\rm 3}
}
\affiliations {
    \textsuperscript{\rm 1}The University of Sydney, Australia\\
    \textsuperscript{\rm 2}Data61, CSIRO, Australia\\
    \textsuperscript{\rm 3}Peking University, China\\
    jie.yin@sydney.edu.au, ke.sun@data61.csiro.au, han.wu@pku.edu.cn
}

\usepackage{bibentry}
\nocopyright

\begin{document}

\maketitle

\begin{abstract}

Graph continual learning (GCL) aims to learn from a continuous sequence of graph-based tasks. Regularization methods are vital for preventing catastrophic forgetting in GCL, particularly in the challenging replay-free, class-incremental setting, where each task consists of a set of unique classes. In this work, we first establish a general regularization framework for GCL based on the curved parameter space induced by the Fisher information matrix (FIM). We show that the dominant Elastic Weight Consolidation (EWC) and its variants are a special case within this framework, using a diagonal approximation of the empirical FIM based on parameters from previous tasks. To overcome their limitations, we propose a new unbiased online curvature approximation of the full FIM based on the model's current learning state. Our method directly estimates the regularization term in an online manner without explicitly evaluating and storing the FIM itself. This enables the model to better capture the loss landscape during learning new tasks while retaining the knowledge learned from previous tasks. Extensive experiments on three graph datasets demonstrate that our method significantly outperforms existing regularization-based methods, achieving a superior trade-off between stability (retaining old knowledge) and plasticity (acquiring new knowledge).

\end{abstract}

\section{Introduction}

Continual learning (CL), also known as lifelong learning or incremental learning, enables models to learn sequentially from a series of tasks to acquire new knowledge. A fundamental challenge inherent to CL is the plasticity-stability dilemma: a crucial trade-off between adapting to new information (plasticity) and preserving previously acquired knowledge (stability). The most prominent consequence is catastrophic forgetting, where a model, upon learning new tasks, suffers from rapid performance degradation on tasks learned earlier. This challenge is coupled with the issue of loss of plasticity (sometimes termed intransigence), indicating the model's diminished capacity to effectively learn new tasks. These difficulties are rooted in the general optimization procedures of neural networks. During training on a specific task, model parameters are optimized to map the task's input distribution to its target outputs. When learning a new task, gradient-based optimization methods, without any constraints, would naturally adjust the learned parameters to minimize the loss with respect to the new task, often %inadvertently
overwriting the knowledge critical for previous tasks. 

To address these inherent challenges, several lines of research have been explored in the last decade. These broadly include: (i) \textit{replay-based methods}, which store and replay old data (or synthetically generated proxy data) from a memory buffer~\citep{caccia2020online,chrysakis2020online,knoblauch2020optimal,lopez2017GEM,shin2017continual}; (ii) \textit{architecture-based methods}, which dynamically increase network capacity or isolate parameters for new tasks~\citep{rusu2016progressive,wortsman2020supermasks,wu2019large,yoon2019scalable,yoon2018lifelong}; and (iii) \textit{regularization-based methods}, which impose constraints on parameter updates to preserve knowledge crucial for previous tasks ~\citep{aljundi18MAS,kirk17,li2017LwF,ritter2018online,saha2020gradient,zenke17SI}. While each offers advantages, they also present trade-offs, such as memory overhead for replay, potentially unbounded model growth for architecture-based  approaches, or known limitations of regularization methods in balancing stability and plasticity in class-incremental settings~\cite{vandeven2022three}.

In this paper, we investigate the 
challenging \textbf{replay-free}, \textbf{class-incremental} setting of graph continual learning (GCL), where a model with a fixed capacity must learn from a new set of classes in each task without replaying from previous tasks. Although recent GCL research has predominantly focused on replay-based methods, which store and replay raw graph nodes~\cite{zhou2021ERGNN}, subgraphs~\cite{zhang2022SSM}, or class prototypes~\cite{niu2024replay}, regularization-based GCL remains a significantly under-explored area. Our work directly address this critical gap. Our contributions are as follows. First, we establish a general regularization framework for GCL grounded in the geometry of the parameter space induced by the Fisher information matrix (FIM). We provide a principled way to estimate parameter importance and constrain their updates, balancing the tradeoff between stability (retaining old knowledge) and plasticity (acquiring new knowledge). We show that the widely used Elastic Weight Consolidation (EWC)~\cite{kirk17} and its variants are a special case of our proposed framework, by taking a diagonal approximation of the empirical FIM based on parameters from previous tasks. Second, we propose a novel method that performs an unbiased online approximation of the FIM based on the current learning state, allowing the model to more accurately capture the loss landscape during learning a new task. Our proposed method directly estimates the FIM-based regularizer in an online manner without the need to explicitly evaluate and store the FIM itself. Third, through extensive experiments on three large-scale graph datasets, we demonstrate that our method achieves a more effective balance between stability and plasticity, outperforming existing regularization techniques.

\section{Related Work}
%\subsection{Continual Learning for Neural Networks}
Continual learning originated in the field of computer vision to address catastrophic forgetting, where a neural network forgets previously learned knowledge upon training on new tasks. CL methods can be broadly grouped into three categories.
\textit{Replay-based methods} address catastrophic forgetting by allowing the model to revisit data information from previous tasks stored in a memory buffer. This is achieved by storing a subset of raw data (experience replay)~\citep{aljundi2019gradient,caccia2020online,chrysakis2020online,knoblauch2020optimal,lopez2017GEM}, or data representations from a generative model (generative replay)~\citep{shin2017continual}. The loss on replayed data is either directly added to the loss on the current data or used as a constraint to guide gradient updates~\cite{lopez2017GEM}. 
\textit{Architecture-based methods} dynamically modify the model's architecture to accommodate the learning of new tasks. They typically expand the network by adding new units or modules for each new task ~\citep{rusu2016progressive,wortsman2020supermasks,wu2019large,yoon2019scalable,yoon2018lifelong}. This ensures that the knowledge from previous tasks is stored in dedicated parts of the model, preventing it from being overwritten.  %and is not overwritten when new tasks are introduced, 
\textit{Regularization-based methods} introduce a regularizer to the loss during learning a new task, which penalizes significant changes to parameters deemed important for previous tasks. %~\citep{aljundi18MAS,farajtabar2020orthogonal,kirk17,li2017LwF,ritter2018online,saha2020gradient,zenke17SI}. 
These methods differ in how they estimate parameter importance. 

Elastic Weight Consolidation (EWC)~\cite{kirk17} is a well-known regularization-based method that estimates the importance of parameters for previous tasks using a diagonal approximation of the empirical FIM. While the original paper does not specify implementation details for computing the diagonal FIM, subsequent works have adopted different implementations~\cite{vandeven2025computationfisher}. For example, some methods compute the squared gradients for each sample's true label with respect to old parameters~\cite{chaudhry2019riemannian}, while others use a randomly sampled label to provide a better approximation~\cite{kao2021natural,vandeven2022three}. Despite its prominence, EWC has several limitations. First, by taking a diagonal approximation of the empirical FIM, it ignores important correlations between parameters. %, leading to suboptimal preservation of previously acquired knowledge. 
Second, EWC requires to store a separate FIM for each previous task, resulting in memory overhead. Third, EWC calculates its diagonal FIM based on the model's parameters from a previous task, which might not be the optimal point when drifts occur during the learning of new tasks.

To improve memory efficiency and scalability, EWC++~\citep{chaudhry2019riemannian} and online EWC~\citep{schwarz2018progress} maintains a decaying moving average of the diagonal FIM as training over tasks progresses, instead of calculating a separate FIM for each task. However, they still rely on a diagonal approximation of the empirical FIM to remain computationally tractable. Synaptic Intelligence (SI)~\cite{zenke17SI} estimates parameter importance during training on a task by accumulating a per-parameter contribution to the decrease in the loss function and uses a first-order approximation to obtain the importance. It is not directly FIM-based but shares the spirit of importance weighting. 
Memory Aware Synapses (MAS)~\cite{aljundi18MAS} estimates parameter importance by measuring the sensitivity of the output function (the gradients of the squared $l_2$ norm of the output) to changes in a parameter.  
%As pointed out in~\cite{benzing22},  SI~\cite{zenke17SI} and MAS~\cite{aljundi18MAS} are shown to be closely linked to EWC as they provide an approximation of the square root of the empirical FIM to measure parameter importance. 
Similar to EWC, both methods also compute per-parameter importance and overlook the correlations between parameters. 
LwF~\cite{li2017LwF} is a functional regularization method that encourages the outputs of the new network to approximate those of the old one. A major limitation of LwF is its efficacy is diminished when tasks are dissimilar, making it inapplicable to class-incremental setting. Other studies~\cite{ritter2018online} use more structured approximations of the FIM (or the Hessian) rather than just the diagonal, such as Kronecker-factored Approximate Curvature (K-FAC), which captures parameter interactions within a same network layer via a block-diagonal approximation of the FIM. However, it leads to significant computational overheads in terms of calculating and inverting two factor matrices and intensive memory cost. While K-FAC is used in optimization~\cite{ritter2018scalable}, its direct application to CL involving many tasks is limited due to these overheads.

\subsubsection{Graph Continual Learning}
Graph continual learning (GCL) is a growing field that seeks to learn from a sequence of graph-based tasks (e.g., node classification). Replay-based methods are the most dominant strategy in GCL, using a memory buffer to revisit historical data when learning a new task. These methods differ in what they store and replay in the memory buffer. 
ERGNN~\cite{zhou2021ERGNN} stores a selection of representative nodes from previous tasks. SSM~\cite{zhang2022SSM} replays sparsified subgraphs to better preserve structural information of previous graphs. To mitigate data imbalance between old and new tasks, Cat~\cite{liu2023CaT} condenses each new graph into a smaller, structurally informative graph before adding it to the memory buffer. For privacy concerns, recent work like TPP~\cite{niu2024replay} has shifted to replaying abstract class prototypes instead of raw graph data. 
%stores and replays prototypes for each learned class, which serve as compact vector representations that encapsulate essential features of a class from past tasks. These prototypes are used to first predict the graph task ID, and then a graph prompting approach is used for within-task node classification. 

Architecture-based methods continually alter the model's architecture to accommodate new graph information. For example, FGN~\citep{wang2022FGN} modifies the GCN architecture to handle the increasing number of nodes in a streaming graph. PI-GNN~\citep{zhang2023PIGNN} expands model parameters to learn new patterns within unstable subgraphs while preventing stable parameters from being rewritten. 

In contrast, regularization-based GCL remains a relatively under-explored area. Most efforts directly apply classic CL regularization methods like  EWC~\cite{kirk17} and LwF~\cite{li2017LwF} without careful scrutiny~\cite{zhang2022CGLB}. TWP~\cite{liu2021TWP} extends EWC by further penalizing updates to parameters important for preserving the topological structure learned from previous tasks. However, its use of globl gradients for calculating parameter importance fails to capture the localized topological information. Our work focuses on developing a novel principled regularization framework for GCL. It can be orthogonally combined with methods from other GCL categories to further mitigate forgetting.

\section{Problem Formulation}
We denote an undirected graph by $\calG = (\calV, \calE, \bm{X})$, where $\calV$ denotes a set of nodes, and $\calE$ denotes a set of edges that connect pairs of nodes. 
$\bm{X} \in \mathbb{R}^{\vert\calV\vert \times d}$ denotes the node feature matrix, where each node $v$ is associated with a feature vector $\bm{x}_v \in \mathbb{R}^d$. Let $\bm{A} \in \mathbb{R}^{\vert\mathcal{V}\vert \times \vert\mathcal{V}\vert}$ denote the graph adjacent matrix, where $\bm{A}(i,j) \in \{0, 1\}$ and $\tilde{\bm{A}}$ denote a normalized graph adjacency matrix with self-loops.

Graph continual learning (GCL) addresses the problem of learning from a sequence of continually incoming graph-based tasks $\{\calT_1, \calT_2, \dots, \calT_T\}$.
Each task $\calT_t = (\calG_t, \calY_t)$ in the sequence is defined as a semi-supervised node classification task, where $\calG_t = (\calV_t, \calE_t, \bm{X}_t)$ denotes a newly emerging subgraph with the node set $\calV_t$, the edge set $\calE_t$, and the node feature matrix $\bm{X}_t \in \mathbb{R}^{|\calV_t| \times d}$. Formally, each task is associated with a set of labeled training nodes $\calV_t^\text{train}$ and a set of unlabeled test nodes $\calV_t^\text{test}$. Each training node $v \in \calV_t^\text{train}$ has a corresponding class label $y_v \in \calY_t$, where $\calY_t$ is the set of classes for task $\calT_t$. %and $\bm{y}_v$ is the one-hot encoding of length $K$.
We consider the class-incremental learning setting, where the class sets of different tasks are disjoint, i.e., $\calY_t \cap \calY_{t'} = \emptyset$ for all $t \neq t'$. The model must learn to classify nodes from a growing set of total classes $\calY = \bigcup_{t=1}^T \calY_t$.
%The training set and the test set for each task are assumed to follow the same distribution. 
The goal is to train a single continual learning model $\calM$ parameterized by $\bm{\theta}$ that can perform node classification on all tasks seen so far. At each step $t$, the model $\calM_t$ is updated sequentially during learning the current task $\calT_t$. %The model $\mathcal{M}_t$ has access only to the data related to the current task, with no or limited access to training data from previous tasks. 
After training on task $\calT_t$, the performance of the model is evaluated on the test sets from all observed tasks up to the point $t$: $\bigcup_{i=1}^{t} \calV_i^\text{test}$. In this work, we focus on the challenging setting of class-incremental learning in a fixed-capacity model, without relying on replaying raw graph data or representations from previous tasks. 

\section{FIM-based Regularization Framework}

Consider a graph convolutional network (GCN)~\cite{kipf2017semi} as the backbone for node classification, its network architecture is specified by the following layer-wise equations: 
\begin{align}
\bm{X}^0 = [\bm{X}\;\bm{1}];\quad&\bm{X}^{l} = [\sigma\left(\bm{H}^l \right)\; \bm{1}]\quad(l=1,\cdots,L-1);\label{eq:gcn}\nonumber\\
\bm{H}^{l+1} &= \tilde{\bm{A}} \bm{X}^{l} \tildewl,\nonumber
\end{align}
where $\bm{X}$ is the input node feature matrix, $\bm{1}$ is a vector of 1's, 
$\bm{H}^l$ is the pre-activation in the $l$'th layer,
$\sigma$ is a non-linear activation function,
$\bm{X}^l$ is the extended activation, which is the activation of the hidden layer outputs augmented by the scalar 1, and $\widetilde{\bm{W}}^l$ is the concatenation of the weight matrix $\bm{W}^l$ with the bias vector $\bm{b}^l$. Given the input features $\bm{X}$ and the corresponding labels $\bm{Y}$, the conditional likelihood is given by
%\begin{equation}
$p( \bm{Y} \mid \bm{X}, \bm{A}, \bm\theta )
=\mathrm{Categorical}(\softmax(\bm{H}^L))$,
%\nonumber\\\end{equation}
where $\bm\theta = \{\widetilde{\bm{W}}^l\}_{l=1}^{L-1}$ denotes all learnable parameters of the GCN, $\softmax(\bm{H}^L)$ applies softmax row-wise to the last layer's pre-activation $\bm{H}^L$, $\mathrm{Categorical}(\cdot)$ means categorical probability masses given by each row of the parameter matrix. For a node classification task, the model is trained by maximizing the conditional likelihood, which is equivalent to minimizing a loss function $\ell(\bm\theta)$ between the predicted probabilities and the true labels.

We formulate a general regularization framework for continual learning, which aims to minimize the regularized loss on the LHS of
\begin{equation}\label{eq:fdivreg}
\ell(\bm\theta)
+
\lambda \sum_{t=1}^{\tau - 1} D_f( \bm\theta_t; \bm\theta )
\approx
\ell(\bm\theta)
+ \lambda \sum_{t=1}^{\tau - 1}
(\bm\theta_t-\bm\theta)^\T
\fim( \bar{\bm\theta} )
(\bm\theta_t-\bm\theta),
\end{equation}
where $\ell$ is the classification loss (in our case the cross entropy loss),
$\lambda>0$ is the regularization strength,
and $D_f$ is the $f$-divergence between two GCN parameter distributions, $\bm\theta_t$ from a previous task and $\bm\theta$ locally at the current task. Under regularity conditions~\cite{localdiv},
regardless of the choice of $D_f$, the regularizer can be written as the RHS,
where $\bar{\bm\theta}$ is a point in between $\bm\theta$ and $\bm\theta_t$,
and $\fim(\bm\theta)$ is the FIM given by
\begin{equation}\label{eq:fim}
\fim_{ij}(\bm\theta)
\defeq
\expect_p \left(
\frac{\partial\Lambda}{\partial\theta_i}
\frac{\partial\Lambda}{\partial\theta_j}
\right)
=\sum_{v\in\calV}
\expect_p \left(
\frac{\partial\Lambda_v}{\partial\theta_i}
\frac{\partial\Lambda_v}{\partial\theta_j}
\right),
\end{equation}
with $\Lambda(\bm\theta)\defeq\log{}p(\bm{Y} \mid \bm{X}, \bm{A}, \bm\theta)$
and
$\Lambda_v(\bm\theta)\defeq\log{}p(\bm{y}_v \mid \bm{X}, \bm{A}, \bm\theta)$.
%\begin{align*}
%p(\bm{X}, \bm{Y}, \bm{A}\mid\bm\theta)
%&=
%p(\bm{X}, \bm{A}) p(\bm{Y} \mid \bm{X},\bm{A}, \bm\theta)\\
%&=
%p(\bm{X}, \bm{A}) \prod_{v\in\calV}p(\bm{y}_v \mid \bm{X},\bm{A}, \bm\theta).
%\end{align*}
%$\bm{A}$ is an adjacency matrix determined by the sampling strategy of neighbourhood (e.g. uniform sampling).
%In the case of GCN, $\bm{A}$ is a constant matrix which can be pre-computed. Note that
%By simple derivations, the FIM is given by
%\begin{equation}\label{eq:fim} \fim_{ij}(\bm\theta) = \end{equation}
%where 
%$\Lambda_v=\log p(\bm{y}_v \mid \bm{X}, \bm{A}, \bm\theta)$ is the conditional log-likelihood, which is
%the same as the negative of the cross entropy loss.
The FIM $\fim(\bm\theta)$ measures the local curvature of the log-likelihood 
$\Lambda$ with respect to infinitesimal variations in the parameter space.
For all $v\in\mathcal{V}$, $\bm{y}_v$ and $\Lambda_v$ only depend on a subgraph 
$\calR_v$ that is the receptive field of node $v$,
and $\bm{y}_v$ is conditional independent given all input features $\bm{X}$ and the graph $\bm{A}$. 
The approximation in \cref{eq:fdivreg} is only accurate when $\bm\theta$ and $\bm\theta_t$ are close enough. In practice, $\bm\theta$ and $\bm\theta_t$ are distant especially for old tasks learned long time ago, and higher order terms of $D_f$ come into play making the approximation inaccurate.

\subsection{Elastic Weight Consolidation (EWC)}

In \cref{eq:fdivreg}, choosing $\bar{\bm\theta} = \bm\theta_t$ and taking the diagonal of the empirical FIM leads to Elastic
Weight Consolidation (EWC) \cite{kirk17, Husz18}. EWC mitigates catastrophic forgetting by selectively penalizing changes to parameters that are important to previous tasks. This is achieved by adding a quadratic regularizer to the loss function of the current task, where the penalty for changing each parameter is weighted by its corresponding diagonal element of the empirical FIM estimated from a previous task.

In EWC, the diagonal of the empirical FIM, denoted as $\fim_{ii}(\bm\theta)$, is calculated and stored after completing each task and before starting a new task. A common method is to compute the expected squared gradients using each node's true label with respect to parameters $\bm\theta_t$:
\begin{equation}\label{eq:ewc}
\fim_{ii}(\bm\theta)
=
\sum_{v\in\calV_t}
\left(
\left.
\frac{\partial\Lambda_v}{\partial\theta_i}\right\vert_{\bm\theta = \bm\theta_t}
\right)^2.
\tag{EWC}
\end{equation}
This method is advocated by \citet{chaudhry2019riemannian} to implement EWC for computational efficiency. An alternative is to use a label randomly sampled from predicted probabilities to compute the squared gradients with respect to the parameters $\bm\theta_t$~\cite{Liu2018rotate,kao2021natural}. 

% For example, the online Laplace approximation uses a block-diagonal (rather than diagonal) approximation \cite{ritter18}.

%\subsection{Our Proposed Method: Motivation}
\subsection{Motivation}

It is important to investigate the rank of the FIM $\fim(\bm\theta)$, which quantifies
the amount of information captured by the classifier $\bm\theta$:
Specifically, it measures how many independent ways of local parameter variations can lead to different models.
\begin{proposition}\label{thm:rank}
Given $\bm{X}$ and $\bm{A}$, $\rank(\fim(\bm\theta))$ is monotonically non-decreasing as $\cal{V}$ grows.
\end{proposition}
%See supplementary material for proofs of our formal statements.
In consequence, as new tasks are learned, $\rank(\fim(\bm\theta))$ with respect to all nodes that the classifier has seen is monotonic, meaning that the classifier captures more and more information to distinguish nodes of different classes --- which makes intuitive sense.

Based on our regularization framework,
we propose to set $\bar{\bm\theta} = \bm\theta$, 
i.e., the model parameters that are being updated in the current task,
and correspondingly, to use $\fim(\bm\theta)$ to measure the squared distance between
$\bm\theta$ and $\bm\theta_t$ in \cref{eq:fdivreg}.
This is because $\bm\theta$ is more recently learned and $\fim(\bm\theta)$ has higher rank by \cref{thm:rank} --- it thus incorporates more information, enabling a more accurate modeling of the regularizer.

Choosing $\bar{\bm\theta}=\bm\theta$ is more reasonable through theoretical lens of information geometry~\cite{aIGI}.
Consider a simplified scenario of class-incremental learning, where the first task is associated with classes 1 and 2, and the second task is associated with classes 3 and 4.
The prediction $p(\bm{y}\mid\bm{x},\bm\theta_1)$ after learning the first task is either class 1 or class 2. 
The learning result denoted as $\bm\theta_1$ is on the edge of the probability simplex $\{(p_1,p_2,p_3,p_4)\}$, where 
$p_1$, $p_2$, $p_3$ and $p_4$ denote probabilities of class 1 to 4, respectively.
%\begin{equation}
%\mathcal{P} = \{
%(p_1,p_2,p_3,p_4)\;:\; p_1+p_2+p_3+p_4=1; \forall{i}, p_i\ge0 \},\nonumber
%\end{equation}
%where $p_3=p_4=0$. 
During training the second task, the parameter $\bm\theta$ has information regarding all 4 classes and is therefore \emph{inside} the probability simplex.
To compute the distance between $\bm\theta$ and $\bm\theta_1$, it is more reasonable to use $\fim(\bm\theta)$ (FIM at the interior point) which is less singular and more informative.

This is also due to the settings of continual learning: we have access to more information regarding the loss landscape at the current task rather than from the historical tasks, where training nodes and graph structure are no longer accessible.
We can therefore more accurately model $\fim(\bm\theta)$ and consider its non-diagonal entries.

%The FIM at $\bm\theta$ has the general expression for continual learning, where we can better approximate FIM with non-diagonal entries. 

Besides its theoretical advantages, our method offers practical benefits. Unlike EWC, which requires storing historial FIMs from previous tasks, we only need to store the learned parameter $\bm\theta_t$ from each previous task.
Practically, we recompute the FIM $\fim(\bm\theta)$ in an \emph{online} manner based on the most recent $\bm\theta$ in the current task. 

\section{Our Proposed Method}
%\jy{Introduce sampling-based GCN and SGD here?}\ke{one or two sentence is sufficient, and may be put along side with the "implementation" paragraph below}

%\subsection{Layer-wise FIM for GCN}
%\subsection{Computational Algorithm}

As it is costly to compute the sum over $v\in\calV$ with a full-batch training of GCNs on large graphs, our method is built upon sampling-based GCNs~\cite{hamilton2017inductive}, which are optimized using stochastic gradient descent (SGD). For each batch $\calB$, the FIM can be approximated by the random matrix:
%In \cref{eq:batch-fim}, we can take the random approximation
$\hat{\fim}_{ij}^{\calB}(\bm\theta)
=
\sum_{v\in\calB}
\left(
\frac{\partial\hat{\Lambda}_v}{\partial\theta_i}
\frac{\partial\hat{\Lambda}_v}{\partial\theta_j}
\right)$,
%\begin{equation}
%\hat{\fim}_{ij}^{\calB}(\bm\theta)
%=
%\sum_{v\in\calB}
%\frac{\partial\hat{\Lambda}_v}{\partial\theta_i}
%\frac{\partial\hat{\Lambda}_v}{\partial\theta_j},
%\end{equation}
where $\hat{\Lambda}_v = \log p(\hat{y}_v \mid \bm{X}, \bm{A}, \bm\theta)$ is the log-likelihood of $\hat{y}_v\sim p(y_v \mid \bm{X}, \bm{A}, \bm\theta)$, a random sample of node $v$'s label based on the current predictive model and node features ($\hat{y}_v$ is independent to node $v$'s given label), and $\calB$ is a random batch of nodes.
It is straightforward to verify that 
$\expect \left(\fim_{ij}^{\calB}(\bm\theta)\right) =\fim_{ij}(\bm\theta)$ and the random matrix gives an unbiased estimate of the FIM.
If the batch size $\vert \calB \vert$ is large enough, $\hat{\fim}^{\calB}(\bm\theta)$ is close to $\fim(\bm\theta)$. 
Using this approximation, we propose to directly compute the regularizer without explicitly evaluating the FIM:\vspace{-3pt}
\begin{align}\label{eq:reg3}
\mathcal{R}(\bm\theta)
&=
\frac{\lambda}{2} \sum_{t=1}^{\tau-1}
(\bm\theta - \bm\theta_t)^\T
\hat{\fim}^{\calB}(\bm\theta)
(\bm\theta - \bm\theta_t),\nonumber\\
%&=
%\frac{1}{2}
%(\bm\theta - \bm\theta_{t-1})^\T
%\sum_{v\in\calB}
%\frac{\partial\hat{\Lambda}_v}{\partial\bm\theta}
%\frac{\partial\hat{\Lambda}_v}{\partial\bm\theta^\T}
%(\bm\theta - \bm\theta_{t-1})\nonumber\\
&=
\frac{\lambda}{2}
\sum_{v\in\calB} \sum_{t=1}^{\tau-1}
\left( \left(
\frac{\partial\hat{\Lambda}_v}{\partial\bm\theta}
\right)^\T
(\bm\theta - \bm\theta_t)\right)^2. \tag{Ours}
\end{align}\vspace{-3pt}
The regularizer is \emph{not} quadratic because 
$\frac{\partial\hat{\Lambda}_v}{\partial\bm\theta}$ is a function of $\bm\theta$. 
The gradient of 
$\frac{\partial\hat{\Lambda}_v}{\partial\bm\theta}$ 
depends on the Hessian of $\hat{\Lambda}_v$.
However, for the ease of implementation, we can consider
$\frac{\partial\hat{\Lambda}_v}{\partial\bm\theta}$
at the current learning step, denoted as $\bm\theta=\bm\theta_{\tau}$, as constant and still apply a quadratic regularizer.

Hence, the loss of the current batch $\calB$ is
$\ell_{\calB}(\bm\theta) = 
\sum_{v\in\calB} \ell_v(\bm\theta) + \mathcal{R}(\bm\theta)$,
where $\ell_v$ is the classification loss of node $v$.
%\begin{equation*}
%\ell_{\calB}
%=
%\sum_{v\in\calB}
%\left[
%\ell_v
%+
%\frac{\lambda}{2}
%\sum_{t=1}^{\tau-1}
%\left( \left(
%\left.
%\frac{\partial\hat{\Lambda}_v}{\partial\bm\theta} \right\vert_{\bm\theta=\bm\theta_{\tau}}
%\right)^\T
%(\bm\theta - \bm\theta_{t})
%\right)^2
%\right],
%\end{equation*}
By a slight abuse of notation, we use $\bm\theta_{1}, \cdots, \bm\theta_{\tau-1}$ to denote the learned parameter values from previous tasks,
but use $\bm\theta_{\tau}$ to denote the parameter that is still being updated in the current task.
By the definition of $\calR$, the gradient of $\ell_{\calB}$ with respect to $\bm\theta$ is
\begin{align}\label{eq:grademc3}
\grad_{\bm\theta}
&=
\sum_{v\in\calB}
\bigg[
\frac{\partial\ell_v}{\partial\bm\theta}\nonumber\\
&+\lambda
\sum_{t=1}^{\tau-1}
\left(
\left.
\frac{\partial\hat{\Lambda}_v}{\partial\bm\theta} \right\vert_{\bm\theta=\bm\theta_{\tau}}
\right)^\T
(\bm\theta - \bm\theta_t)
\left.
\frac{\partial\hat{\Lambda}_v}{\partial\bm\theta} \right\vert_{\bm\theta=\bm\theta_{\tau}}
\bigg],
\end{align}
which is usually evaluated at $\bm\theta=\bm\theta_{\tau}$, the current parameter value, depending on the optimizer.
%$\bm\theta_{\tau}$ is the value of $\bm\theta$ at the current learning step in task $\tau$, 

\begin{proposition}
The regularizer $\mathcal{R}$ is an unbiased estimate of the regularizer (the second term) on the RHS of \cref{eq:fdivreg} up to scaling.
\end{proposition}
\begin{remark}
As an unbiased estimate, $\mathcal{R}$ utilizes the \emph{full FIM} $\fim(\bm\theta)$ rather than its (block-)diagonal approximations.
This is a key theoretical advantage over EWC, which is limited to the diagonal FIM.
The computational and storage complexity of the full FIM is effectively circumvented based on its low-rank random approximation.
\end{remark}

% below are striaghtforward and revmoed for clarity
%Therefore the loss of the current batch $\calB$ is
%\begin{equation*}
%\ell_{\calB}
%=
%\sum_{v\in\calB}
%\left[
%\Lambda_v
%+
%\frac{\lambda}{2}
%\left( \left(
%\left.
%\frac{\partial\hat{\Lambda}_v}{\partial\bm\theta} \right\vert_{\bm\theta=\bm\theta_{\tau}}
%\right)^\T
%(\bm\theta - \bm\theta_{t-1})
%\right)^2
%\right],
%\end{equation*}
%where $\bm\theta_{\tau}$ is the value of $\bm\theta$ at the current learning step $\tau$ in the current task, and $\lambda>0$ is the regularization strength.
%Let $\bm\theta_{\tau}$ denote the parameter that is still getting updated in the current task.
%Its gradient w.r.t. $\bm\theta$ is
%\begin{equation}\label{eq:grademc3}
%\grad_{\bm\theta}
%=
%\sum_{v\in\calB}
%\left[
%\frac{\partial\Lambda_v}{\partial\bm\theta}
%+
%\lambda
%\left(
%\left.
%\frac{\partial\hat{\Lambda}_v}{\partial\bm\theta} \right\vert_{\bm\theta=\bm\theta_{\tau}}
%\right)^\T
%(\bm\theta - \bm\theta_{t-1})
%\left.
%\frac{\partial\hat{\Lambda}_v}{\partial\bm\theta} \right\vert_{\bm\theta=\bm\theta_{\tau}}
%\right],
%\end{equation}
%which is usually evaluated at $\bm\theta=\bm\theta_{\tau}$, the current parameter value, depending on the optimizer.

The proposed regularizer makes intuitive sense.
Observe that the gradient is a linear combination of 
$\frac{\partial\Lambda_v}{\partial\bm\theta}$
and
$\frac{\partial\hat{\Lambda}_v}{\partial\bm\theta}$,
where
$\frac{\partial\Lambda_v}{\partial\bm\theta}$
is the gradient of the nodes in the batch $\calB$,
and
$\frac{\partial\hat{\Lambda}_v}{\partial\bm\theta}$ is the gradient of node $v\in\calB$ with an ``imaginary'' label $\hat{y}_v$ (which could be different from $y_v$ for multi-modal prediction associated with ambiguous nodes). Therefore, the regularization essentially augments the current batch of nodes with imaginary labels. One advantage of this $\grad_{\bm\theta}$ is that it fully respects the gradient of an augmented batch of nodes, which may not be true for alternative regularizers.

The weights of these augmented nodes can be distinct for different $v\in\calB$.
There are two following cases:
\begin{enumerate}
\item[1)]
$$
\left( \left. \frac{\partial\hat{\Lambda}_v}{\partial\bm\theta} \right\vert_{\bm\theta=\bm\theta_{\tau}}
\right)^\T
(\bm\theta_{\tau} - \bm\theta_t) \ge 0,
$$ which means the negative gradient w.r.t. the imaginary sample 
$
-
\left. \frac{\partial\hat{\Lambda}_v}{\partial\bm\theta} \right\vert_{\bm\theta=\bm\theta_{\tau}}
$
has an accurate angle with $\bm\theta_t-\bm\theta_{\tau}$, the vector pointing from $\bm\theta_{\tau}$ to the previous task $t$. The coefficient of $\frac{\partial\hat{\Lambda}_v}{\partial\bm\theta}$ is positive, meaning that the imaginary sample contributes positively to the gradient computation, and learning encourages to going towards the previous parameter value $\bm\theta_t$;
\item[2)]
$$
\left( \left. \frac{\partial\hat{\Lambda}_v}{\partial\bm\theta} \right\vert_{\bm\theta=\bm\theta_{\tau}}
\right)^\T
(\bm\theta_{\tau} - \bm\theta_t) < 0.$$
In this case, the coefficient of $\frac{\partial\hat{\Lambda}_v}{\partial\bm\theta}$ is negative, meaning that the imaginary sample is a negative sample.
Learning discourages to going away from the previous parameter value $\bm\theta_t$. This makes intuitive sense.
\end{enumerate}

\subsection{Algorithm Implementation} 
For implementation, we first perform a forward pass of the current batch of nodes to compute the classification loss
$\ell'_{\calB}=\sum_{v\in\calB}\ell_v$,
which is the sum of the per-sample loss. We then run a backward pass to calculate the gradient of the loss $\frac{\partial\ell'_\calB}{\partial\bm\theta}
=
\sum_{v\in\calB}
\frac{\partial\Lambda_v}{\partial\bm\theta}
$. To compute the gradient of the regularizer, we sample a label
$\hat{y}_v\sim\mathrm{Multinomial}(\mathrm{softmax}(\bm{c}_v^L)$ independently for each $v\in\calB$, where $\bm{c}^L_v$ is the output logits before applying softmax. Then, we run a second backward pass to compute the gradient 
$\frac{\partial\hat{\Lambda}_v}{\partial\bm\theta}$,
which is a $\vert \calB \vert \times \dim(\bm\theta) $ matrix, or the Jacobian matrix of the mapping $\bm\theta\to\bm{\hat{\Lambda}}_v$ (the vector by concatenating all $\hat{\Lambda}_v$ for $v\in\calB$), given by
\begin{align}
\frac{\partial\Lambda''_{\calB}}{\partial\bm\theta}
&=
\sum_{v\in\calB} 
\left[ \sum_{t=1}^{\tau-1}
\left(
\frac{\partial\hat{\Lambda}_v}{\partial\bm\theta}
\right)^\T 
(\bm\theta-\bm\theta_t)
\right]
\cdot
\frac{\partial\hat{\Lambda}_v}{\partial\bm\theta}\nonumber,
\end{align}
where we get $\vert\calB\vert$ coefficients to linearly combine the rows of the matrix 
$\frac{\partial\hat{\Lambda}_v}{\partial\bm\theta}$.
Finally, the gradient of the loss plus the regularizer is
\begin{equation}
\grad_{\bm\theta}
=
\frac{\partial\ell'_{\calB}}{\partial\bm\theta}
+
\lambda
\frac{\partial\Lambda''_{\calB}}{\partial\bm\theta}\nonumber.
\end{equation}
The above gradient $\grad_{\bm\theta}$ can then be fed into the optimizer to update model parameters $\bm\theta$. Therefore, two backward passes are required to compute the gradient of the regularized loss. 

\begin{algorithm}[t]  
  \caption{Unbiased Online Curvature Approximation}  
  \label{alg:training}
  \textbf{Input}: A sequence of tasks $\calT=\{\calT_1, \ldots, \calT_T\}$, a queue $\calQ$ of size $M$, learning rate $\alpha$, hyperparameter $\beta$\\
  \textbf{Output}: A GNN $f_{\bm\theta_T}(\cdot)$ %GCN $\{\bm\theta_1, \ldots, \bm\theta_T\}$
  \begin{algorithmic}[1] %[1] enables line numbers
  \FOR{$t = 1, \ldots, T$}
    \STATE $\calQ = \emptyset$,
    \FOR{epoch = $1, \ldots, K$}
    \IF{$t = 1$}
    \STATE \texttt{// initial training on $\calT_1$}
    \STATE Compute the loss $\ell_\calB = \sum_{v \in \calB} \ell_v$
    \ELSE
    \STATE \texttt{// Incremental training on $\calT_{t>1}$} 
    \STATE Randomly sample one node $v \in \calB$
    \STATE Randomly sample $\hat{y}_v$ and compute $\hat\Lambda_v$
    \STATE Compute the gradient $\frac{\partial\hat{\Lambda}_v} {\partial\bm\theta}$
    \IF{$\vert\calQ\vert > M$} 
        \STATE $\calQ.\text{pop}()$
    \ENDIF
    \STATE $\calQ.\text{push}(\frac{\partial\hat{\Lambda}_v} {\partial\bm\theta})$     
    \STATE Compute the regularized loss $\ell_\calB$ using \cref{eq:batchloss}
    \ENDIF
    \STATE $\ell_\calB.\text{backward}()$
    \STATE $\bm\theta \leftarrow \bm\theta - \alpha \frac{\partial\ell_\calB} {\partial\bm\theta}$
    \ENDFOR
    \STATE \texttt{// exponential moving average}
    \STATE $\bm{\theta}_t \leftarrow \beta \bm{\theta}_{t-1} + (1-\beta) \bm{\theta}$
  \ENDFOR
  %\STATE \textbf{return} A GNN $f_{\bm\theta_T}(\cdot)$
\end{algorithmic}
\end{algorithm}

The primary computational overhead is the additional backward pass required to compute the gradient of the regularizer  $\frac{\partial \hat{\Lambda}_v}{\partial \bm{\theta}}$ for each batch. To improve efficiency, we propose a gradient caching strategy, which maintains a queue $\calQ$ to store $M$ gradient vectors, representing $\frac{\partial \hat{\Lambda}_v}{\partial{\bm\theta}}$ in the recent optimization steps (``near'' the current $\bm\theta$). The queue is initialized as empty at the start of each new task and is  updated throughout the model's training process. For each batch $\calB$, we first perform a forward pass for all nodes in the batch. We then randomly sample a node $v\in\calB$ and accordingly sample a label $\hat{y}$. A backward pass is thereafter executed to compute the gradient $\frac{\partial \hat{\Lambda}_v}{\partial \bm{\theta}}$. This new gradient is pushed to the end of the queue. If the queue reaches its maximum size $M$, the front element is popped before enqueuing the new gradient vector. This method allows the model to reuse previously computed gradients stored in the queue across batches and avoid a full second backward pass on every batch, thus significantly improving computational efficiency. %The memory overhead is \(\mathcal{O}(M \times \dim(\bm{\theta}))\), where \(M\) is the queue size and \(\dim(\bm{\theta})\) is the dimensionality of the model parameters.
%$\mathcal{O}(M\times\dim(\bm{x})\times\dim(\bm{h}))$.
% More concretely, for a two-layer GCN, the queue can be represented as
% \begin{equation}
% \left\{ (\partial\hat{\Lambda}_v/\partial\bm{W}_1,
% \partial\hat{\Lambda}_v/\partial\bm{b}_1,
% \partial\hat{\Lambda}_v/\partial\bm{W}_2,
% \partial\hat{\Lambda}_v/\partial\bm{b}_2) \right\}_{v\in\calQ}.
% \end{equation}

\begin{table*}[t]
\centering
\caption{Performance comparison with baseline methods on CoraFull, Arxiv, and Coauthor-CS.}
\setlength{\tabcolsep}{8pt}
\scalebox{1}{
\begin{tabular}{c|cc|cc|cc}
    \toprule
    \multirow{2}{*}{\textbf{Methods}} & \multicolumn{2}{c|}{\textbf{CoraFull}} & \multicolumn{2}{c|}{\textbf{Arxiv}} & \multicolumn{2}{c}{\textbf{Coauthor-CS}} \\
     & AP$\uparrow$ & AF$\uparrow$ & AP$\uparrow$ & AF$\uparrow$ & AP$\uparrow$ & AF$\uparrow$ \\
    \midrule
    EWC$_\text{empirical}$   & 18.19 & -86.91 & 11.01  & -72.44 & 19.84 & -95.72 \\
    EWC$_\text{sample}$ & 17.78 & -85.02 &  12.48 & -76.88 & 20.05 & -95.29 \\
    EWC$_\text{pred}$& 18.10 & -84.89 &  12.21  & -75.89 & 23.64 &-91.13 \\\midrule
    Online EWC ($\gamma = 1$) & 18.07 & -84.68 &  12.50  & -74.52 & 19.72 & -95.92 \\
    Online EWC ($\gamma = 0.95$) & 17.80 & -85.16 &  12.30  & -76.02 & 19.69 & -95.94\\
    MAS   & 18.27 & -84.01 &  10.81 & -70.16 & 46.30 & -61.73 \\
    LwF   & 18.38  & -87.53  & 11.67  &   -76.31 & 31.92 & -80.11 \\
    TWP   & 18.04 & -87.27 & 11.06  & -72.54 & 21.26 & -93.80 \\
    %GEM   & 17.95  & -86.85 & 11.74 & -74.50 & 30.29 & -82.12 \\   \midrule
    %ERGNN & 17.97  & -86.63 & 12.37  & -78.41& 92.50 & -4.35 \\
    \midrule
    Ours  & 46.93 & -8.70 &  12.46 & -1.94 & 34.55 & -60.18 \\
    Ours (w/o EMA)   & 45.59 & -43.01 & 11.63 & -6.04 & 27.16 & -78.2 \\
    \bottomrule
\end{tabular}
}
\label{tab:comparison}
\end{table*}

Based on the GCN and gradient vectors stored in $\calQ$, the total loss for each batch can be computed as:
\begin{align}
\label{eq:batchloss}
\ell_{\calB}
&=
\sum_{v\in\calB} \ell_v
+
\frac{\lambda}{2 \vert \calQ \vert}\sum_{v\in\calQ}
\left[ \sum_{t=1}^{\tau-1}
\left( \frac{\partial\hat{\Lambda}_v}{\partial\bm\theta} \right)^\T
( \bm\theta - \bm\theta_{t-1} )
\right]^2.\vspace{-3pt}
% &
% =
% \sum_{v\in\calB} \ell_v
% +
% \frac{\lambda}{2 \vert \calQ \vert}\sum_{v\in\calQ}
% \Bigg[ \sum_{t=1}^{\tau-1}
% \biggr[
% \left( \frac{\partial\hat{\Lambda}_v}{\partial\bm{W}_1} \circ ( \bm{W}_1 - {\bm{W}_1}_t) \right).\mathrm{sum}()
% +
% \left( \frac{\partial\hat{\Lambda}_v}{\partial\bm{b}_1} \circ ( \bm{b}_1 - {\bm{b}_1}_t ) \right).\mathrm{sum}()
% \\
% &+
% \left( \frac{\partial\hat{\Lambda}_v}{\partial\bm{W}_2} \circ ( \bm{W}_2 - {\bm{W}_2}_t) \right).\mathrm{sum}()
% +
% \left( \frac{\partial\hat{\Lambda}_v}{\partial\bm{W}_2} \circ ( \bm{W}_2 - {\bm{W}_2}_t) \right).\mathrm{sum}()
% \bigg]
% \Bigg]^2.
\end{align}
Then, the gradients $\frac{\partial \hat{\Lambda}_v}{\partial \bm{\theta}}$ are obtained via $\ell_{\calB}.\text{backward}()$, and fed to the optimizer to update model parameters. 

After completing training on each task, we also apply an exponential moving average (EMA) to further mitigate catastrophic forgetting. The parameters are updated as: 
$$\bm{\theta}_t \leftarrow \beta \bm{\theta}_{t-1} + (1-\beta) \bm{\theta},$$
where $\bm{\theta}$ denotes the newly learned parameters for the current task, $\bm\theta_{t-1}$ denotes the parameters stored from the previous task, and $\beta$ is a hyperparameter. The resulting parameter $\bm{\theta}_t$ is stored to initialize the next task. The training procedure of our proposed algorithm in summarized in Algorithm~\ref{alg:training}.

\section{Experiments}

\subsection{Experimental Setup}
\subsubsection{Datasets}
Three large graph datasets are used: CoraFull~\cite{kachites2000Cora}, Arxiv\footnote{\url{https://ogb.stanford.edu/docs/nodeprop/#ogbn-arxiv}}~\cite{hu2020OGB}, and Coauthor-CS~\cite{shchur2018CS}, for the task of node classification. %On CoraFull, following TWP~\cite{liu2021TWP}, we retain classes with more than 150 nodes and construct five tasks with nine classes per task. On Arxiv, we retain classes with more than 410 nodes and construct six tasks, each task containing six classes of nodes. For Coauthor-CS, we retain all classes and construct five tasks with three classes per task.
For all datasets, the model is trained sequentially on all tasks and evaluated on test sets for all learned tasks. For each class, we allocate 60\% of the nodes for training, 20\% for validation, and 20\% for testing. Please see supplementary material for task configuration tasks. 

%The details of the three datasets and task configuration are summarized in Table~\ref{tab:benchmarks}.

\begin{table}[h]
    \centering
    \caption{Statistics of three benchmark datasets.}\vspace{-4pt}
    \label{tab:benchmarks}
    \setlength{\tabcolsep}{1mm}
    \begin{tabular}{c|ccccc}
    \toprule
    Datasets   & \textbf{CoraFull} &  \textbf{Arxiv} & \textbf{Coauthor-CS} \\
    \midrule
    \# Nodes   & 17,407    &  168,391  &    18,333\\
    \# Edges   & 92,221   &  972,962 &   154,837 \\
    \# Classes & 45       &  36      &   15     \\
    \# Tasks   & 5        &  6       &   5\\
    \# Classes per task & 9 & 6 &   3\\
    Avg \# nodes per task & 3,481 & 28,065    &3,667\\
    Avg \# edges per task &  18,444 & 162,160 & 30,967\\
    \bottomrule
    \end{tabular}
\end{table}

\subsubsection{Baselines} For a fair comparison, we mainly compare our method against other regularization-based CL methods. All baselines operate under the same setting: using a fixed-capacity model without any form of replay. The detailed descriptions of these baselines are as follows:

\begin{itemize}
    \item \textbf{EWC}~\cite{kirk17}: We compare three implementations of EWC, which differ in how the loglikelihood is calculated: $\textbf{EWC}_\text{empirical}$ uses each node's true label~\cite{Liu2018rotate},  $\textbf{EWC}_\text{sample}$ uses a label randomly sampled from predicted probabilities~\cite{kao2021natural,vandeven2022three}, and $\textbf{EWC}_\text{pred}$ uses the predicted label~\cite{vandeven2025computationfisher}.
    \item \textbf{Online EWC}~\cite{schwarz2018progress,chaudhry2019riemannian} as an additional baseline. Unlike EWC~\cite{kirk17}, which stores a separate FIM for each previous task, Online EWC maintains a single decaying moving average of the diagonal FIM as training progresses over tasks. This process is controlled by a decay rate hyperparameter $\gamma$, where a larger value of $\gamma$ places more weight on the FIM from previous tasks.
    
    \item \textbf{MAS}~\cite{aljundi18MAS} measures the importance of parameters by the sensitivity of the learned function output to a parameter change.
    
    \item \textbf{LwF}~\cite{li2017LwF} is minimizes the discrepancy between the logits of the old and new models.
    
    %\item \textbf{GEM}~\cite{lopez2017GEM} is a replay-based method that stores a subset of samples in the episodic memory and modifies the gradients of the current task by projecting it into a subspace that does not conflict with those calculated from the memory. We include GEM as a baseline as it also incorporates a form of regularization to constrain gradient updates.
    
    \item \textbf{TWP}~\cite{liu2021TWP} is a regularization method tailored to graph data that penalizes updates to parameters critical for preserving topological structure. 
\end{itemize}

\subsubsection{Evaluation Metrics} For evaluation, we use two widely used metrics, average performance (AP) and average forgetting (AF)~\citep{lopez2017GEM}, measuring the plasticity and stability of the model. The definitions are given as follows:
\begin{equation}
    \textrm{AP}_t = \frac{\sum_{i=1}^t M_{t,i}}{t}, \quad \textrm{AF}_t = \frac{\sum_{i=1}^{t-1} (M_{t,i}- M_{i,i})}{t-1}, \nonumber
\end{equation}\vspace{-2pt}
where $M_{t,i} $ denotes the prediction accuracy on task $\mathcal{T}_i$ after the model has been trained on task $\calT_t  \ (t=1, \ldots, T)$. For both metrics, the higher the better. Due to the inherent trade-off between AP and AF, both metrics should be taken into account in performance evaluation. 

\begin{figure*}[t]
     \centering
     \begin{subfigure}[b]{\textwidth}
         \centering
         \includegraphics[width=\textwidth]{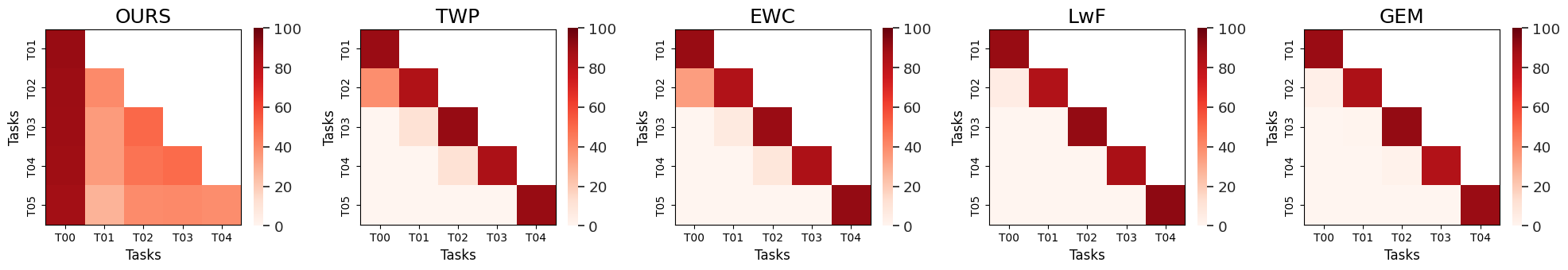}
         \caption{CoraFull}
         \label{fig:ewc}
        \end{subfigure}\\
        \begin{subfigure}[b]{\textwidth}
         \centering
         \includegraphics[width=\textwidth]{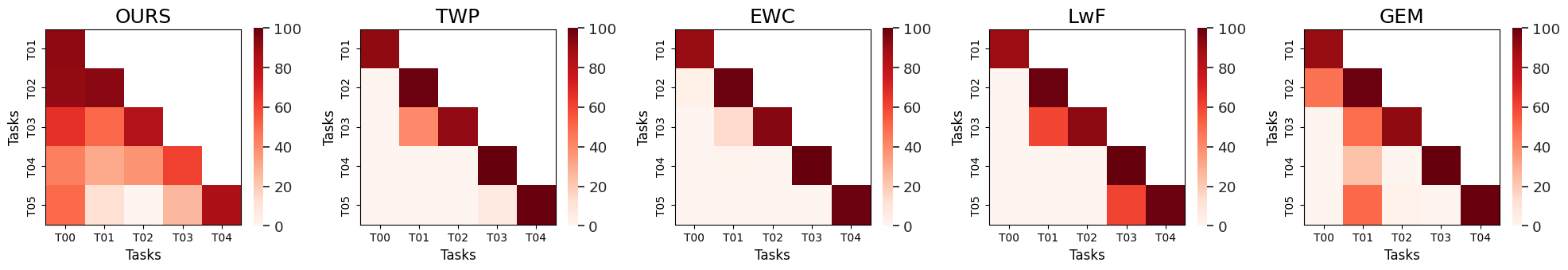}
         \caption{Author-CS}
         \label{fig:lwf}
        \end{subfigure}
        \caption{Performance heatmaps of the final model evaluated on all previous tasks for (a) CoraFull and (b) Author-CS.}
        \label{fig:heatmaps}
\end{figure*}

\subsubsection{Implementation Details} 
For all methods, we use a two-layer GCN as the backbone, with a hidden dimension of 256. 
Following the settings reported in the original papers, the regularization strength $\lambda$ is set to $10^9$, while $\lambda_1$ and $\lambda_2$ are both set to $10^4$ with $\beta$ set to 0.01 for TWP. %For ER-GNN, the budget is specified as $[100, 1000]$ with a depth $d$ of 0.5. 
For LwF, the lambda distance $\lambda_{\text{dist}}$ is set to a range of $[1.0, 10.0]$, and temperature parameter $T$ is specified as a range of $[2.0, 20.0]$. EWC and MAS both uses a memory strength of $10^4$. %while GEM uses a memory strength of 0.5 and retains a fixed number of 100 memories. 
For Online EWC, We report the results using two $\gamma$ values: $\gamma = 1$ and $\gamma = 0.95$ (the value suggested in the original work~\cite{schwarz2018progress}).
The regularization strength $\lambda$ is set to $10^4$, the same as for EWC. These hyperparameters were carefully chosen to match the configurations from the respective original methods, ensuring consistency across experiments. 

For our method, the hyperparameter $\beta$ for EMA is set to 0.5. Adam is used as the optimizer with a learning rate of 0.00001 with a weight decay $5e-4$ and a batch size of 128. The queue size $M$ is tuned in $\{32, 64, 128\}$ and the optimal value is found to be 128.  
All of the reported results are averaged over 3 runs with different random seeds. All models are implemented in PyTorch and trained on a single Tesla V100 GPU.

% \begin{figure*}[th!]
%     \centering
%     \includegraphics[width=\textwidth]{figs/combined_plot.png}
%     \caption{Performance heatmaps of the final model evaluated on all previous tasks.}
%     \label{fig:heatmaps}
% \end{figure*}

\subsection{Main Results and Analyses}

In Table~\ref{tab:comparison}, we compare our proposed method with existing regularization-based CL methods for node classification on three graph datasets. Our method outperforms three EWC variants by a large margin, especially on CoraFull and Coauthor-CS. This affirms that our method provides a more accurate approximation of the full FIM compared to EWC's diagonal approximation of the empirical FIM. Our method also significantly surpasses Online EWC on all datasets in terms of both AP and AF. We attribute this substantial performance gap to the limitations of the diagonal FIM approximation inherited by Online EWC, which fails to account for parameter correlations.
Consequently, Online EWC is insufficient to accurately capture the true loss landscape, resulting in severe catastrophic forgetting despite its online design. These empirical results support the theoretical advantage of our method in providing a more accurate and unbiased online FIM approximation (Proposition 2), thereby more effectively mitigating forgetting in a continual learning setting. 

Furthermore, while TWP was specifically designed for graph data, its performance is comparable to traditional CL methods and substantially worse than our method. The ablation of the exponential moving average (\textbf{Ours (w/o EMA)} leads a modest drop on AP but a significant drop on AF, highlighting its crucial role in mitigating catastrophic forgetting.

Figure~\ref{fig:heatmaps} presents heatmap plots to compare the performance of a final model evaluated on all previous tasks on CoraFull and Coauthor-CS. These plots reveal two key observations: (1) The results clearly show the limitations of EWC. As seen across all datasets, EWC achieves near-zero accuracy on most previous tasks, indicating its ineffectiveness in mitigating catastrophic forgetting. This poor performance is due to its simplified approximation of the FIM, failing to adequately capture the true importance of model parameters. This finding resonates with the conclusions from ~\cite{vandeven2022three}, which noted the inapplicability of existing regularization-based CL methods to class-incremental  setting. (2) Our method demonstrates a significantly better ability to handle catastrophic forgetting. Built upon its theoretical advantages, our approach empirically retains high performance on previous tasks while maintaining strong accuracy on the current task. This indicates a more effective stability-plasticity balance under the challenging class-incremental setting, which is a significant improvement over existing regularization-based CL methods.

\subsection{Effect of Regularization Strength}

\begin{figure}[h]
     \centering
     \begin{subfigure}[b]{0.42\textwidth}
         \centering
         \includegraphics[width=\textwidth]{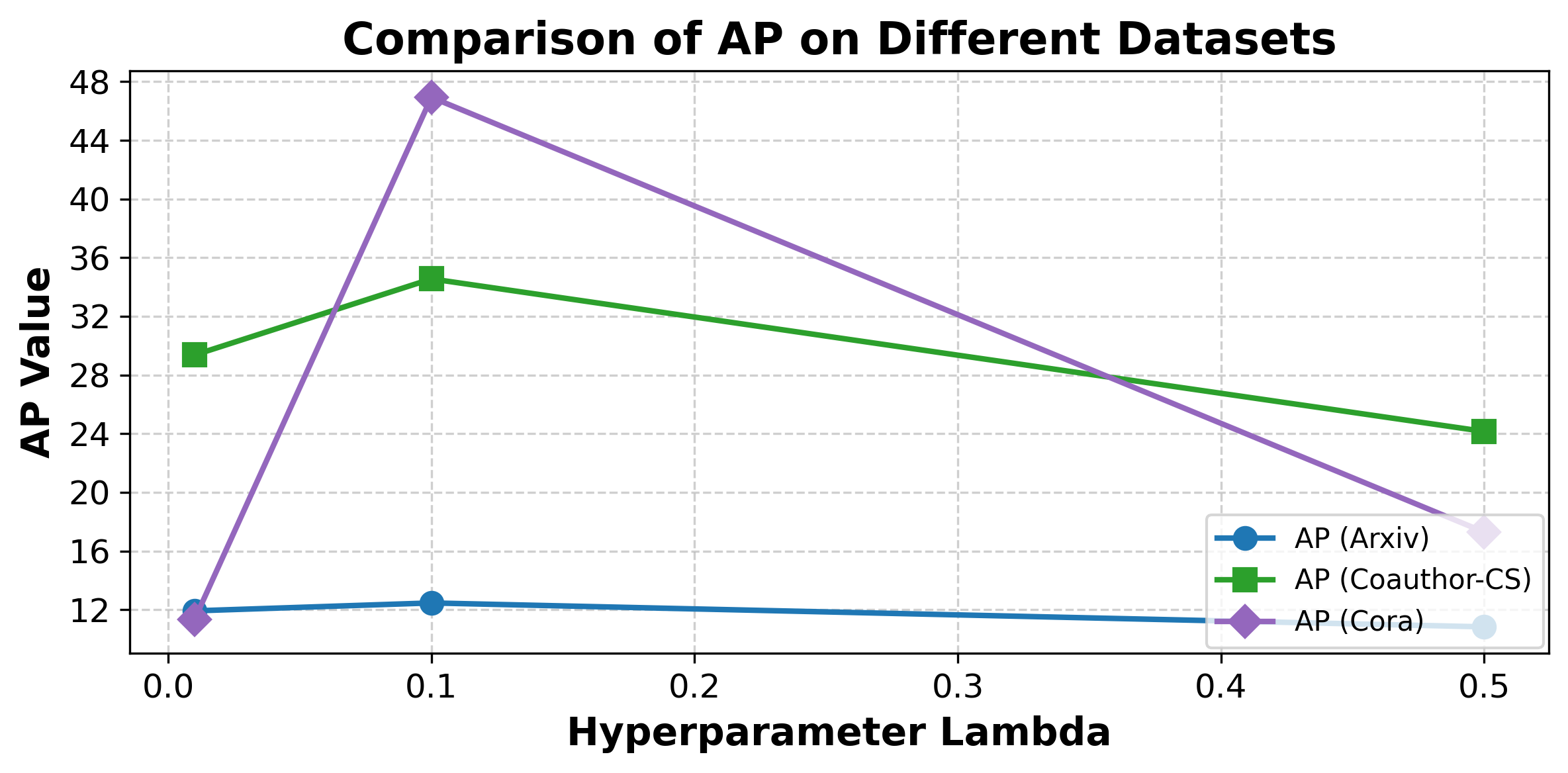}
         % \caption{AP}
         \label{fig:AP}
        \end{subfigure}\\
        \vspace{-0.1cm}
        \begin{subfigure}[b]{0.42\textwidth}
         \centering
         \includegraphics[width=\textwidth]{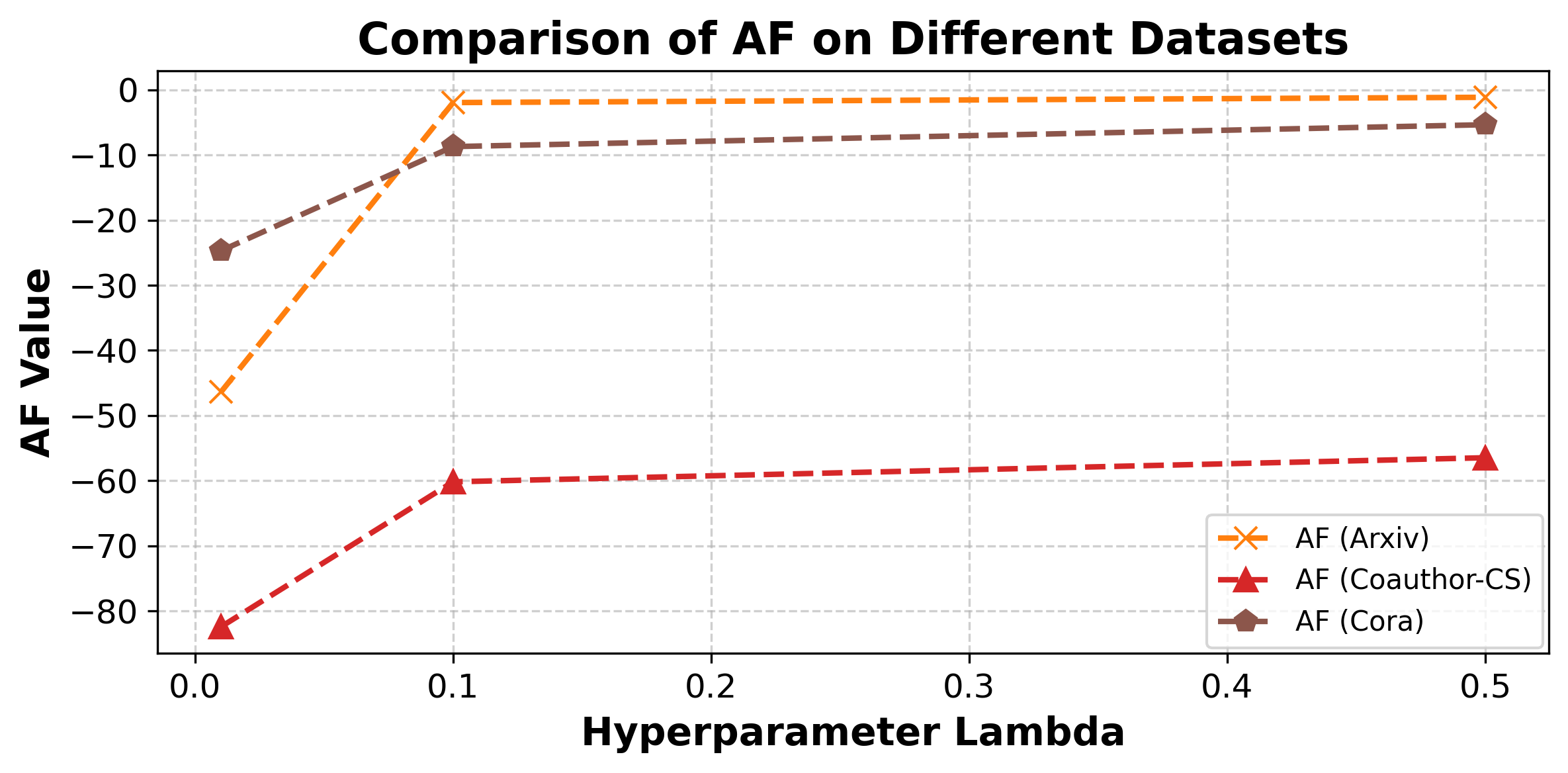}
         % \caption{AF}
         \label{fig:AF}
         \vspace{-0.3cm}
        \end{subfigure}
        \caption{Effect of regularization strength $\lambda$.}
        \label{fig:lambda}
        \vspace{-3pt}
\end{figure}

Figure~\ref{fig:lambda} shows the performance of our method with varying regularization strengths $\lambda$. In contrast to methods like EWC and TWP, which typically require a large regularization strength, our method performs optimally with a smaller value for $\lambda$. 
we tune the values of $\lambda$ within $\{0.01, 0.1, 0.5\}$, and find that our method consistently achieves the best AP and AF when $\lambda = 0.1$ on all three datasets. 

\section{Conclusion}
We established a general FIM-based regularization framework for graph continual learning, addressing the challenging replay-free, class-incremental setting with a fixed model capacity. We theoretically show that our method improves upon classic regularization methods like EWC by using an unbiased, online FIM approximation that more accurately models the loss landscape without the memory overhead of storing historical FIMs. Experiments on three graph datasets demonstrated that our method %achieves a more effective balance between stability and plasticity, 
outperforms existing regularization-based CL methods. Future work will investigate how our regularization method could be integrated with GCL methods from other categories.

\bibliography{gcl}

\begin{thebibliography}{40}
\providecommand{\natexlab}[1]{#1}

\bibitem[{Aljundi et~al.(2018)Aljundi, Babiloni, Elhoseiny, Rohrbach, and
  Tuytelaars}]{aljundi18MAS}
Aljundi, R.; Babiloni, F.; Elhoseiny, M.; Rohrbach, M.; and Tuytelaars, T.
  2018.
\newblock Memory Aware Synapses: Learning What (not) to Forget.
\newblock In \emph{ECCV}, 144--161.

\bibitem[{Aljundi et~al.(2019)Aljundi, Lin, Goujaud, and
  Bengio}]{aljundi2019gradient}
Aljundi, R.; Lin, M.; Goujaud, B.; and Bengio, Y. 2019.
\newblock Gradient based Sample Selection for Online Continual Learning.
\newblock In \emph{NeurIPS}, volume~32, 11816--11825.

\bibitem[{Amari(2016)}]{aIGI}
Amari, S. 2016.
\newblock \emph{Information Geometry and Its Applications}, volume 194 of
  \emph{Applied Mathematical Sciences}.
\newblock Springer-Verlag, Berlin.

\bibitem[{Andrew Kachites~McCallum and Seymore(2000)}]{kachites2000Cora}
Andrew Kachites~McCallum, J.~R., Kamal~Nigam; and Seymore, K. 2000.
\newblock Automating the construction of internet portals with machine
  learning.
\newblock volume~3, 127--163.

\bibitem[{Blyth(1994)}]{localdiv}
Blyth, S. 1994.
\newblock Local Divergence and Association.
\newblock \emph{Biometrika}, 81(3): 579--584.

\bibitem[{Caccia et~al.(2020)Caccia, Belilovsky, Caccia, and
  Pineau}]{caccia2020online}
Caccia, L.; Belilovsky, E.; Caccia, M.; and Pineau, J. 2020.
\newblock Online Learned Continual Compression with Adaptive Quantization
  Modules.
\newblock In \emph{ICML}, 1240--1250. PMLR.

\bibitem[{Chaudhry et~al.(2019)Chaudhry, Dokania, Ajanthan, and
  Torr}]{chaudhry2019riemannian}
Chaudhry, A.; Dokania, P.~K.; Ajanthan, T.; and Torr, P. H.~S. 2019.
\newblock Riemannian Walk for Incremental Learning: Understanding Forgetting
  and Intransigence.
\newblock In \emph{ECCV}, 532--547.

\bibitem[{Chrysakis and Moens(2020)}]{chrysakis2020online}
Chrysakis, A.; and Moens, M.-F. 2020.
\newblock Online Continual Learning from Imbalanced Data.
\newblock In \emph{ICML}, 1952--1961. PMLR.

\bibitem[{Hamilton, Ying, and Leskovec(2017)}]{hamilton2017inductive}
Hamilton, W.~L.; Ying, R.; and Leskovec, J. 2017.
\newblock Inductive Representation Learning on Large Graphs.
\newblock In \emph{NeurIPS}, 1025--1035.

\bibitem[{Huszár(2018)}]{Husz18}
Huszár, F. 2018.
\newblock On Quadratic Penalties in Elastic Weight Consolidation.
\newblock \emph{Proceedings of the National Academy of Sciences}, 115(11).

\bibitem[{Kao et~al.(2021)Kao, Jensen, van~de Ven, Bernacchia, and
  Hennequin}]{kao2021natural}
Kao, T.-C.; Jensen, K.~T.; van~de Ven, G.~M.; Bernacchia, A.; and Hennequin, G.
  2021.
\newblock Natural Continual Learning: Success is a Journey, not (just) a
  Destination.
\newblock In \emph{NeurIPS}, 28067--28079.

\bibitem[{Kipf and Welling(2017)}]{kipf2017semi}
Kipf, T.~N.; and Welling, M. 2017.
\newblock Semi-Supervised Classification with Graph Convolutional Networks.
\newblock In \emph{ICLR}.

\bibitem[{Kirkpatrick et~al.(2017)Kirkpatrick, Pascanu, Rabinowitz, Veness,
  Desjardins, Rusu, Milan, Quan, Ramalho, Grabska-Barwinska, Hassabis, Clopath,
  Kumaran, and Hadsell}]{kirk17}
Kirkpatrick, J.; Pascanu, R.; Rabinowitz, N.; Veness, J.; Desjardins, G.; Rusu,
  A.~A.; Milan, K.; Quan, J.; Ramalho, T.; Grabska-Barwinska, A.; Hassabis, D.;
  Clopath, C.; Kumaran, D.; and Hadsell, R. 2017.
\newblock Overcoming Catastrophic Forgetting in Neural Networks.
\newblock \emph{Proceedings of the National Academy of Sciences}, 114(13):
  3521--3526.

\bibitem[{Knoblauch, Husain, and Diethe(2020)}]{knoblauch2020optimal}
Knoblauch, J.; Husain, H.; and Diethe, T. 2020.
\newblock Optimal Continual Learning has Perfect Memory and is {NP}-hard.
\newblock In \emph{ICML}, 5327--5337. PMLR.

\bibitem[{Li and Hoiem(2018)}]{li2017LwF}
Li, Z.; and Hoiem, D. 2018.
\newblock Learning without Forgetting.

\bibitem[{Liu, Yang, and Wang(2021)}]{liu2021TWP}
Liu, H.; Yang, Y.; and Wang, X. 2021.
\newblock Overcoming Catastrophic Forgetting in Graph Neural Networks.
\newblock In \emph{AAAI}, 8653--8661.

\bibitem[{Liu et~al.(2018)Liu, Masana, Herranz, Van~de Weijer, López, and
  Bagdanov}]{Liu2018rotate}
Liu, X.; Masana, M.; Herranz, L.; Van~de Weijer, J.; López, A.~M.; and
  Bagdanov, A.~D. 2018.
\newblock Rotate your Networks: Better Weight Consolidation and Less
  Catastrophic Forgetting.
\newblock In \emph{ICPR}, 2262--2268.

\bibitem[{Liu, Qiu, and Huang(2023)}]{liu2023CaT}
Liu, Y.; Qiu, R.; and Huang, Z. 2023.
\newblock {CaT}: Balanced Continual Graph Learning with Graph Condensation.
\newblock In \emph{ICDM}, 1157--1162.

\bibitem[{Lopez-Paz and Ranzato(2017)}]{lopez2017GEM}
Lopez-Paz, D.; and Ranzato, M. 2017.
\newblock Gradient Episodic Memory for Continual Learning.
\newblock In \emph{NeurIPS}, 6467--6476.

\bibitem[{Niu et~al.(2024)Niu, Pang, Chen, and Liu}]{niu2024replay}
Niu, C.; Pang, G.; Chen, L.; and Liu, B. 2024.
\newblock Replay-and-forget-free graph class-incremental learning: a task
  profiling and prompting approach.
\newblock In \emph{NeurIPS}, 87978--88002.

\bibitem[{Ritter, Botev, and Barber(2018{\natexlab{a}})}]{ritter2018online}
Ritter, H.; Botev, A.; and Barber, D. 2018{\natexlab{a}}.
\newblock Online Structured Laplace Approximations For Overcoming Catastrophic
  Forgetting.
\newblock In \emph{NeurIPS}, 3742--3752.

\bibitem[{Ritter, Botev, and Barber(2018{\natexlab{b}})}]{ritter2018scalable}
Ritter, H.; Botev, A.; and Barber, D. 2018{\natexlab{b}}.
\newblock A Scalable Laplace Approximation for Neural Networks.
\newblock In \emph{ICLR}.

\bibitem[{Rusu et~al.(2016)Rusu, Rabinowitz, Desjardins, Soyer, Kirkpatrick,
  Kavukcuoglu, Pascanu, and Hadsell}]{rusu2016progressive}
Rusu, A.~A.; Rabinowitz, N.~C.; Desjardins, G.; Soyer, H.; Kirkpatrick, J.;
  Kavukcuoglu, K.; Pascanu, R.; and Hadsell, R. 2016.
\newblock Progressive Neural Networks.
\newblock In \emph{arXiv preprint arXiv:1606.04671}.

\bibitem[{Saha, Garg, and Roy(2021)}]{saha2020gradient}
Saha, G.; Garg, I.; and Roy, K. 2021.
\newblock Gradient Projection Memory for Continual Learning.
\newblock In \emph{ICLR}.

\bibitem[{Schwarz et~al.(2018)Schwarz, Czarnecki, Luketina, Grabska-Barwinska,
  Teh, Pascanu, and Hadsell}]{schwarz2018progress}
Schwarz, J.; Czarnecki, W.~M.; Luketina, J.; Grabska-Barwinska, A.; Teh, Y.~W.;
  Pascanu, R.; and Hadsell, R. 2018.
\newblock Progress \& Compress: A Scalable Framework for Continual Learning.
\newblock In \emph{ICML}, 4528--4537. PMLR.

\bibitem[{Shchur et~al.(2018)Shchur, Mumme, Bojchevski, and
  G{\"{u}}nnemann}]{shchur2018CS}
Shchur, O.; Mumme, M.; Bojchevski, A.; and G{\"{u}}nnemann, S. 2018.
\newblock Pitfalls of Graph Neural Network Evaluation.
\newblock \emph{arXiv preprint}, arXiv: 1811.05868.

\bibitem[{Shin et~al.(2017)Shin, Lee, Kim, and Kim}]{shin2017continual}
Shin, H.; Lee, J.~K.; Kim, J.; and Kim, J. 2017.
\newblock Continual Learning with Deep Generative Replay.
\newblock In \emph{NeurIPS}, volume~30, 2990--2999.

\bibitem[{van~de Ven(2025)}]{vandeven2025computationfisher}
van~de Ven, G.~M. 2025.
\newblock On the Computation of the {F}isher Information in Continual Learning.
\newblock arXiv:2502.11756.

\bibitem[{van~de Ven, Tuytelaars, and Tolias(2022)}]{vandeven2022three}
van~de Ven, G.~M.; Tuytelaars, T.; and Tolias, A.~S. 2022.
\newblock Three Types of Incremental Learning.
\newblock \emph{Nature Machine Intelligence}, 4(11): 1185--1197.

\bibitem[{Wang et~al.(2022)Wang, Qiu, Gao, and Scherer}]{wang2022FGN}
Wang, C.; Qiu, Y.; Gao, D.; and Scherer, S. 2022.
\newblock Lifelong Graph Learning.
\newblock In \emph{CVPR}, 13719--13728.

\bibitem[{Weihua~Hu and Leskovec(2020)}]{hu2020OGB}
Weihua~Hu, M. Z. Y. D. H. R. B. L. M.~C., Matthias~Fey; and Leskovec, J. 2020.
\newblock Open graph benchmark: Datasets for machine learning on graphs.
\newblock In \emph{NeurIPS}, volume~33, 22118--22133.

\bibitem[{Wortsman et~al.(2020)Wortsman, Ramanujan, Liu, Kembhavi, Rastegari,
  Yosinski, and Farhadi}]{wortsman2020supermasks}
Wortsman, M.; Ramanujan, V.; Liu, R.; Kembhavi, A.; Rastegari, M.; Yosinski,
  J.; and Farhadi, A. 2020.
\newblock Supermasks in Superposition.
\newblock volume~33, 15173--15184.

\bibitem[{Wu et~al.(2019)Wu, Chen, Wang, Ye, Liu, Guo, and Fu}]{wu2019large}
Wu, Y.; Chen, Y.; Wang, L.; Ye, Y.; Liu, Z.; Guo, Y.; and Fu, Y. 2019.
\newblock Large Scale Incremental Learning.
\newblock In \emph{CVPR}, 374--382.

\bibitem[{Yoon et~al.(2019)Yoon, Kim, Yang, and Hwang}]{yoon2019scalable}
Yoon, J.; Kim, S.; Yang, E.; and Hwang, S.~J. 2019.
\newblock Scalable and Order-robust Continual Learning with Additive Parameter
  Decomposition.
\newblock In \emph{ICLR}.

\bibitem[{Yoon et~al.(2018)Yoon, Yang, Lee, and Hwang}]{yoon2018lifelong}
Yoon, J.; Yang, E.; Lee, J.; and Hwang, S.~J. 2018.
\newblock Lifelong Learning with Dynamically Expandable Networks.
\newblock In \emph{ICLR}.

\bibitem[{Zenke, Poole, and Ganguli(2017)}]{zenke17SI}
Zenke, F.; Poole, B.; and Ganguli, S. 2017.
\newblock Continual Learning through Synaptic Intelligence.
\newblock In \emph{ICML}, 3987--3995.

\bibitem[{Zhang et~al.(2023)Zhang, Yan, Li, Wang, Xie, Song, and
  Kim}]{zhang2023PIGNN}
Zhang, P.; Yan, Y.; Li, C.; Wang, S.; Xie, X.; Song, G.; and Kim, S. 2023.
\newblock Continual Learning on Dynamic Graphs via Parameter Isolation.
\newblock In \emph{SIGIR}, 601--611. ACM.

\bibitem[{Zhang, Song, and Tao(2022{\natexlab{a}})}]{zhang2022CGLB}
Zhang, X.; Song, D.; and Tao, D. 2022{\natexlab{a}}.
\newblock {CGLB}: Benchmark Tasks for Continual Graph Learning.
\newblock In \emph{NeurIPS}, volume~35, 13006--13021.

\bibitem[{Zhang, Song, and Tao(2022{\natexlab{b}})}]{zhang2022SSM}
Zhang, X.; Song, D.; and Tao, D. 2022{\natexlab{b}}.
\newblock Sparsified Subgraph Memory for Continual Graph Representation
  Learning.
\newblock In \emph{ICDM}, 1335--1340.

\bibitem[{Zhou and Cao(2021)}]{zhou2021ERGNN}
Zhou, F.; and Cao, C. 2021.
\newblock Overcoming Catastrophic Forgetting in Graph Neural Networks with
  Experience Replay.
\newblock In \emph{AAAI}, 4714--4722.

\end{thebibliography}


%\input{ReproducibilityChecklist.tex}

\end{document}